\pdfoutput=1

\documentclass[11pt]{article}

\usepackage[final]{acl}

\usepackage{times}
\usepackage{latexsym}

\usepackage[T1]{fontenc}

\usepackage[utf8]{inputenc}

\usepackage{microtype}

\usepackage{inconsolata}

\usepackage{graphicx}

\usepackage{amsmath}
\usepackage{algpseudocode}
\usepackage{algorithm} 
\usepackage{multirow}
\usepackage{booktabs}
\usepackage{enumitem}

\usepackage{hyperref}

\title{AssoCiAm: A Benchmark for Evaluating \underline{Asso}ciation Thinking while \underline{Ci}rcumventing \underline{Am}biguity}

\author{
 \textbf{Yifan Liu\textsuperscript{1}\thanks{Equal contribution}},
 \textbf{Wenkuan Zhao\textsuperscript{1}\footnotemark[1]},
 \textbf{Shanshan Zhong\textsuperscript{1}},
 \textbf{Jinghui Qin\textsuperscript{2}},
\\
 \textbf{Mingfu Liang\textsuperscript{3}},
 \textbf{Zhongzhan Huang\textsuperscript{1}},
 \textbf{Wushao Wen\textsuperscript{1}\thanks{Corresponding author}},
\\
\\
 \textsuperscript{1}Sun Yat-sen University,
 \textsuperscript{2}Guangdong University of Technology,
 \textsuperscript{3}Northwestern University
}

\begin{document}
\maketitle
\begin{abstract}
Recent advancements in multimodal large language models (MLLMs) have garnered significant attention, offering a promising pathway toward artificial general intelligence (AGI). Among the essential capabilities required for AGI, creativity has emerged as a critical trait for MLLMs, with association serving as its foundation. Association reflects a model's ability to think creatively, making it vital to evaluate and understand. While several frameworks have been proposed to assess associative ability, they often overlook the inherent ambiguity in association tasks, which arises from the divergent nature of associations and undermines the reliability of evaluations. To address this issue, we decompose ambiguity into two types—internal ambiguity and external ambiguity—and introduce AssoCiAm, a benchmark designed to evaluate associative ability while circumventing the ambiguity through a hybrid computational method. We then conduct extensive experiments on MLLMs, revealing a strong positive correlation between cognition and association. Additionally, we observe that the presence of ambiguity in the evaluation process causes MLLMs' behavior to become more random-like. Finally, we validate the effectiveness of our method in ensuring more accurate and reliable evaluations. See \href{https://github.com/lyf15/AssoCiAm}{Project Page} for the data and codes.

\end{abstract}

\section{Introduction}
\label{sec:intro}
Recently, multimodal large language models~(MLLMs) have made remarkable advancements, enabling them to better understand the human world and mimic human multimodal perception~\cite{li2024survey, liang2023toa, liang2024aide}. This progress highlights a potential pathway toward achieving artificial general intelligence~(AGI), a long-term goal in the field of artificial intelligence~\cite{yin2024survey, huang2025minilongbench}. As a hallmark of human intelligence, creativity is a critical ability that AGI-level MLLMs should possess and be evaluated for~\cite{bellemare2024divergent}. A key aspect of evaluating creativity lies in assessing associative ability, which plays a central role in creativity by driving generative processes~\cite{beaty2023associative, benedek2012associative, marron2018chain, benedek2020elements}. 

\begin{figure}[t]
  \includegraphics[width=1\linewidth]{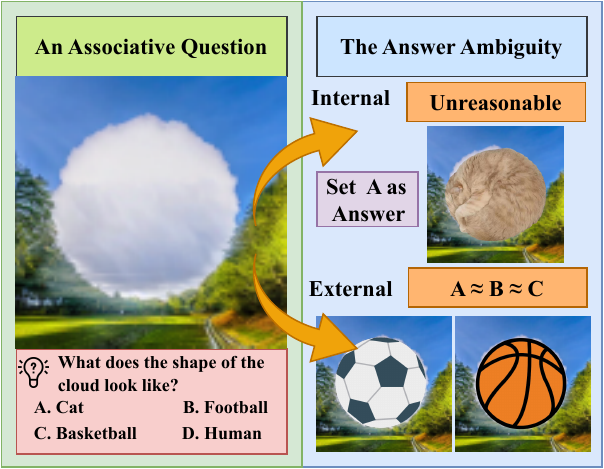} 
  \caption {
    An example illustrating the inherent ambiguity in association evaluation. The left image presents an associative task where models associate the shape of the cloud with the most similar option from the given choices. The right image highlights two types of ambiguity: internal ambiguity (top), where the ground truth is set to option A with the unreasonable explanation that the cat’s curled posture resembles a circle; and external ambiguity (bottom), where options B and C are equally correct due to their similar circular shapes but are not designated as the correct answers in this context.
  }
  \label{fig:intro}
\end{figure}

Associative ability refers to the process of forming novel connections between seemingly unrelated concepts stored in memory~\cite{beaty2023associative}. This process is divergent, originating from a single idea and expanding into a space of possible ideas through paradigms such as divergent thinking, lateral thinking or "thinking outside the box". To evaluate associative abilities, numerous researchers have proposed diverse evaluation frameworks and released various benchmarks, exploring different dimensions of this ability. Most of these benchmarks~\cite{jiang2023brainteaser, kraaijveld2024columbus, lin2021riddlesense, zhang2022birdqa, zhong2024let} adopt a multiple-option format, where a single answer is designated as correct.

However, these benchmarks overlook the answer ambiguity inherent in associative tasks, which stems from the divergent nature of associations. This ambiguity can be explicitly decomposed into two parts: \textbf{internal} and \textbf{external} ambiguity.

(1) \textbf{Internal Ambiguity.}
Internal ambiguity refers to cases where the designated answer itself is unreasonable. For example, in the associative task shown in Fig.~\ref{fig:intro}~(left), the task requires the models to associate the shape of a cloud with an object that resembles it~\cite{zhong2024let}. Here, the correct answer is set as Option A, "cat", because a cat can curl into a circular shape, as shown in Fig.~\ref{fig:intro}~(right). However, this answer is unreasonable since the typical shape of a cat does not match the shape of the cloud. With internal ambiguity, models cannot identify the correct answer from options even if they possess strong associative ability, and therefore the evaluation results fail to reflect the true associative ability of models.

(2) \textbf{External Ambiguity.}
External ambiguity arises when the correct options include multiple equally correct options. As illustrated in Fig.~\ref{fig:intro}~(right), the correct answer, Option A, is not unique; Options B and C are also valid because their circular shapes closely resemble the shape of the cloud. In this associative task, external ambiguity occurs because multiple options share a resemblance to the natural object's shape. In such cases, directly judging a model's answer against a ground truth lead to inaccurate evaluations.

Neglecting internal or external ambiguity introduces risks that render evaluation results non-faithful. Effectively circumventing answer ambiguity is critical for ensuring that benchmarks produce reliable evaluations. To address this issue, we construct a multimodal associative benchmark that circumvents ambiguity and provides more faithful quantification of associative ability. 

In our task design, MLLMs are required to perform a associative task, such as that illustrated in Fig.~\ref{fig:intro}~(left), where they identify the shape of a specific natural object in an image and then associate it with the most visually similar object. Ambiguity in this task arises from divergent visual associations. To mitigate ambiguity, we propose a novel hybrid computational method that mitigates both internal and external ambiguity.

Using our method, we construct the first multimodal associative benchmark circumventing ambiguity, AssoCiAm, to quantify models' associative abilities. Furthermore, we conduct large-scale evaluations on various models and identify a strong correlation between associative ability and cognitive capability. Through a series of experiments, we analyze the impact of ambiguity on association evaluation, which leads to random-like behavior in models, and validate the effectiveness of our method in mitigating ambiguity. Our contributions are summarized as follows:
\begin{itemize}[leftmargin=0.3cm]
\vspace{-0.35cm}
    \item 
    We first introduce the answer ambiguity inherent in association evaluation and discuss its impact on the reliability of association evaluation.
\vspace{-0.35cm}
    \item 
    We construct AssoCiAm, for evaluating associative ability and propose a hybrid computational method to circumvent the ambiguity.
\vspace{-0.35cm}
    \item 
    We conduct extensive experiments to assess associative abilities of MLLMs, show the impact of ambiguity on association evaluation, and validate the effectiveness of our proposed method.

\end{itemize}

\section{Related Work}
\label{sec:rew}
\textbf{Multimodal Large Language Model.}
MLLMs are LLM-based systems capable of receiving, reasoning and producing outputs across multiple modalities~\cite{yin2024survey}. To endow models with such capabilities, researchers typically position pre-trained LLMs~\cite{chung2024scaling, touvron2023llama, vicuna2023, bai2023qwen, touvron2023llama2} as the brain of the system while employing multimodal encoders~\cite{zhao2025accelerating, cherti2023reproducible, radford2021learning, sun2023eva} as sensory components to process multimodal information. Recently, various techniques have been proposed~\cite{li2023mimic, zhong2023adapter, lu2022learn, zhang2023multimodal,huang2025routereval, zhong2024moextend, yang2024gpt4tools, wu2024v} to enhance the emergent capabilities of MLLMs, bridging the gap between human and artificial intelligence. These advancements underscore the 
importance of evaluation in the research and development of MLLMs~\cite{huang2024survey}.

\noindent{\textbf{Computational Creativity.}}
The emergence of increasingly powerful MLLMs has reignited interest in computational creativity~\cite{ismayilzada2024creativity}. Computational creativity encompasses a broad range of tasks, including linguistic creativity~\cite{mittal2022ambipun, arora2022transfer, xie2025funqa, chakrabarty2023spy, ismayilzada2024evaluating}, creative problem solving~\cite{jiayang2023storyanalogy, lewis2024using, opielka2024large, mitchell2023comparing, ahrabian2024curious}, and artistic creativity~\cite{yang2022re3, popescu2023gpoet, shi2023detecting, brooks2024video, copet2024simple}. A key underlying capability for all these forms of creativity is associative ability, as these creative tasks rely on associative thinking to form connections between diverse concepts. Therefore, gaining insight into creativity necessitates a deeper understanding of associative ability, which serves as the core of creativity.

\noindent{\textbf{Associative Thinking Evaluation.}}
Various frameworks have been developed to evaluate associative ability from diverse perspectives. For instance, DAT~\cite{chen2023probing} assesses models' associative abilities by measuring semantic distance, reporting that models outperform humans in this aspect. Additionally, the Alternate Uses Test~\cite{guilford1967nature} and the Torrance Tests of Creative Thinking have been used to evaluate models, with studies indicating that GPT-3~\cite{brown2020language} and GPT-4~\cite{achiam2023gpt} achieve near-human performance~\cite{goes2023pushing, guzik2023originality, hubert2024current, koivisto2023best}. Despite these advancements highlighting progress in associative tasks primarily dependent on divergent thinking, models continue to show gaps compared to humans on associative tasks requiring lateral thinking or "thinking outside the box"~\cite{jiang2023brainteaser, kraaijveld2024columbus, lin2021riddlesense, zhang2022birdqa, zhong2024let,huang2025causality}. Our benchmark specifically focuses on this type of tasks, addressing the unique ambiguities inherent in associative evaluations.

\section{The Need for New Benchmarks}
Current benchmarks have provided valuable insights into the associative abilities of MLLMs while also highlighting their limitations. However, there is still room for improvement. Given the divergent nature of association, the ambiguity present in different datasets vary and often remains implicit. As a result, the ambiguity in an associative benchmark go unnoticed during its design. This makes it essential to explicitly account for ambiguity when constructing benchmarks to ensure more reliable and fair assessments. 
Furthermore, we analyze several existing benchmarks, including BiRdQA~\cite{zhang2022birdqa}, BrainTeaser~\cite{jiang2023brainteaser}, and RiddleSense~\cite{lin2021riddlesense}, by sampling 70 questions from each. Our analysis reveals that 15.7\%, 24.3\%, and 17.1\% of the questions from these benchmarks contain ambiguity, indicating that ambiguity is both prevalent and non-trivial. These findings point to the need for greater fairness in evaluation and highlight the importance of developing new benchmarks that explicitly and systematically mitigate the ambiguity.
 

\section{Construction of AssoCiAm}
\label{sec:cons}

In this section, we present the pipeline for constructing AssoCiAm that mitigates ambiguity.

\subsection{Overview of The Pipeline}

AssoCiAm follows a multiple-option question-answering format.
A test sample consists of two main components. (1) \textbf{An image.} As illustrated in Fig.~\ref{fig:regen}~(d, e, f), an image is a mask, e.g. Fig.~\ref{fig:regen}~(b), filled with natural objects such as clouds, beaches, and waterfalls, embedded in a natural background. (2) \textbf{A Question-Options pair.} Questions and options are presented in text format. For each question, models are given m options, only one of which is correct; we term this an mT1 question. Our benchmark includes three subtasks: 4T1, 7T1, and 10T1, each consisting of a series of mT1 questions with 4, 7, and 10 options, respectively.

To construct the benchmark, we design a two-stage pipeline. (1)~\textbf{Avoiding Internal Ambiguity}, which collects images while avoiding internal ambiguity.
We collect representative masks that eliminate internal ambiguity. Using these masks as guidance, we generate images and review them to ensure quality and consistency.
(2)~\textbf{Avoiding External Ambiguity}, which constructs Question-Options pairs while mitigating external ambiguity.
We introduce a structured way to model the options to systematically mitigate external ambiguity. Based on this model, we apply an optimization method to select appropriate distractors and construct reliable test samples.



\subsection{Avoiding Internal Ambiguity}
\label{sec:filter}
We propose a hybrid computational approach to eliminate the internal ambiguity during the process of collecting masks and images that we need. In this context, eliminating internal ambiguity means ensuring that shapes of masks are representative.

\begin{figure}[htbp]
  \includegraphics[width=\linewidth]{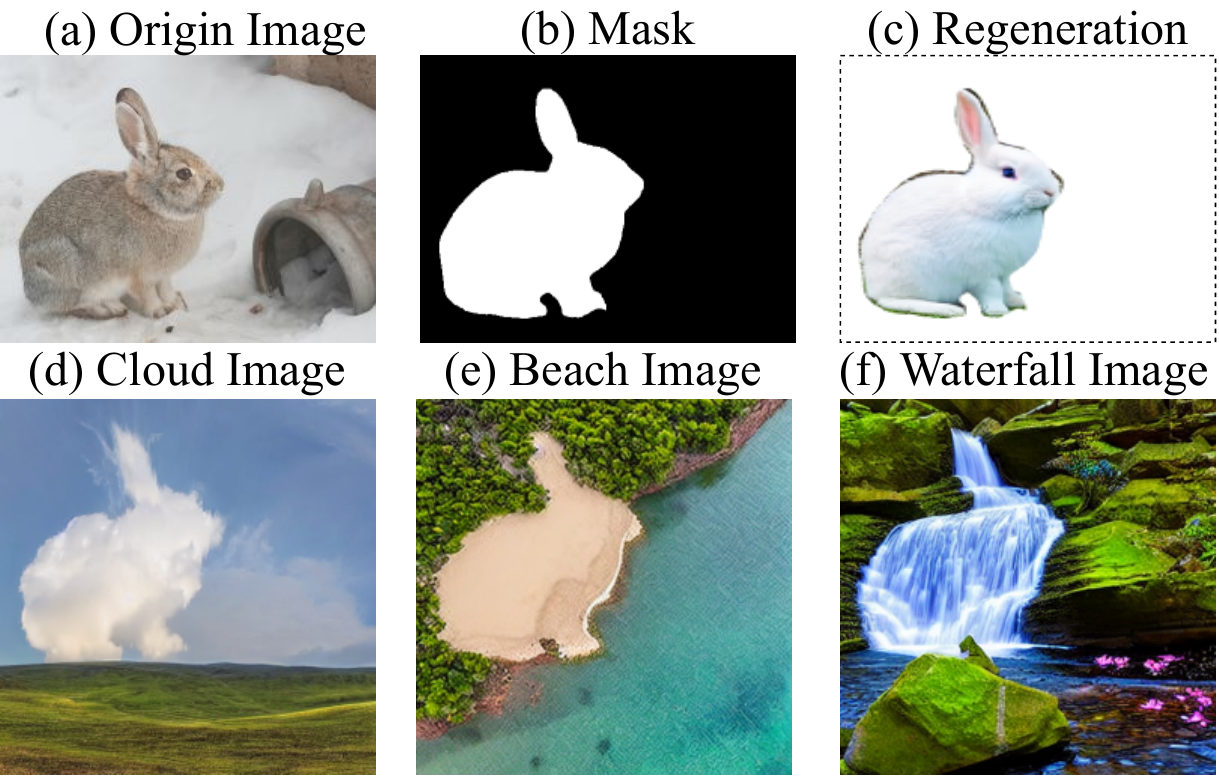} 
  \caption {Examples for images used in our method. (a) is a image collected. (b) is the mask extracted from (a). (c) are images regenerated guided by (b) and overlaid with (b). (d, e, f) are final images used in our benchmark.
  }
    \label{fig:regen}
\end{figure}

First, we collect masks extracted from images. The images come from the publicly available ILSVRC12 dataset~\cite{ILSVRC15}. The dataset is specifically designed for image classification, focusing on typical features of each class, making it well-suited for extracting representative masks.
We sample 25 images per class and use the SAM model~\cite{kirillov2023segany} to extract their corresponding masks. For instance, Fig.~\ref{fig:regen}~(b) is the mask extracted from Fig.~\ref{fig:regen}~(a).

Second, we filter and augment masks that represent image classes, following a \textbf{key principle}: a representative mask should be inherently recognizable and align with general human perception.

To facilitate this, we first apply a computational filtering method to simplify the selection process and help human experts work more efficiently and precisely when refining and selecting representative masks.
The method is based on the premise that a truly representative mask should enable a \textbf{regenerated image}, guided by the mask, to be \textbf{recognized} as belonging to its original class.

Specifically, we use a control diffusion model~\cite{rombach2022high, zhang2023adding, huang2023scalelong, shi2023exploring} to regenerate images guided by masks, with irrelevant regions overlaid to reduce visual distraction. For each mask, eight images are regenerated to mitigate randomness. As shown in Fig.~\ref{fig:regen}~(c), a rabbit image is regenerated and overlaid with its corresponding mask from Fig.~\ref{fig:regen}~(b).


Next, we use CLIP~\cite{radford2021learning} to classify the regenerated images. A mask's representativeness is measured by the average classification probability across its regenerated images. Masks scoring above 97\% are retained for further review.

Human experts then are invited to select the \textbf{most representative} masks from this \textbf{filtered set}. Additionally, human experts include a small number of masks sourced from the public website \href{https://www.flaticon.com/}{flaticon}, which provides free-to-use images for project purposes. These masks, though excluded from the selected classes, are commonly associated with the context of our task. This final collection of masks thus constitutes the \textbf{original mask set}.

With the original mask set, we use the control diffusion model~\cite{rombach2022high, zhang2023adding, huang2023scalelong, shi2023exploring} to generate images of clouds, beaches, and waterfalls, each guided by its corresponding mask, as shown in Fig.~\ref{fig:regen}~(d, e, f). A subsequent manual filtering step ensures consistency, clarity, and completeness of the generated images, thereby eliminating internal ambiguity in AssoCiAm.

\subsection{Avoiding External Ambiguity}
\label{sec:alg}

To construct Question-Options pairs for AssoCiAm, we use the selected classes as candidate distractors. To avoid \textbf{external ambiguity}, it is essential to prevent distractor classes with shapes similar to the correct answer from being selected. Since the most representative mask for each class has been identified, the similarity between shapes of two classes corresponds to the similarity between their corresponding masks. 

Given this, we represent the relationships among options using a graph. Specifically, we define an undirected complete graph $ G = \langle V, E \rangle $, where $ V = \{v_1, v_2, \dots, v_n\} $ represents the set of classes, and $ E = \{e_{ij}\} $ is the set of edges. Each edge $ e_{ij} $ quantifies the similarity between masks $ v_i $ and $ v_j $, serving as a measure of their resemblance.

To determine the optimal set of distractors and answers, we define a subgraph $ G' = \langle V', E' \rangle $, where $ V' \subseteq V $ and $ E' \subseteq E $, with $ v_0 \in V' $ representing the correct answer. 
Ensuring the uniqueness of the correct answer corresponds to minimizing the following function:
\begin{equation}
S(G') = \sum_{v_i \in V', i \neq 0 } \frac{1}{|V'|-1} e_{0i},
\label{eq:init}
\end{equation}
where $ e_{ij} $ represents the similarity between the classes $ v_i  $ and $ v_j $. This ensures that distractors are as dissimilar as possible from the correct answer.


However, minimizing $ S(G') $ alone can lead to a scenario where the distractors are highly similar to each other. This allows the model to identify the correct answer by eliminating the distractors based on their mutual resemblance rather than truly recognizing the correct answer. To address this issue, we include the \textbf{variance} $ \sigma^2(G') $ of all edge weights in $ G' $ as a regularization term, which measures the spread of similarities among options relative to their mean and does not directly indicate the magnitude of similarity. The combined objective function is expressed as:
\begin{equation}
F(G') = S(G') + \lambda \sigma^2(G'),
\label{eq:final}
\end{equation}
where $ \sigma^2(G') $ represents the variance of all the edge weights, and $ \lambda $ is a regularization parameter that controls the trade-off between minimizing similarity and maintaining variance. 


To construct the graph, we calculated the similarity between masks via DINO-v2~\cite{oquab2023dinov2, darcet2023vitneedreg}, which focuses exclusively on their geometric shapes since the masks are binary (black and white) and differ primarily in shape. Then, Genetic Algorithm~\cite{mirjalili2019genetic, holland1992genetic, mitchell1998introduction, goldberg2013genetic} is applied for the optimization. The implementation of the Genetic Algorithm is provided in Appendix~\ref{app:ga}.





Finally, we construct Question-Options pairs for the benchmark within the generated images and the algorithm. For each mT1 subtask, we select three questions per image. These questions express the same meaning but are phrased differently to query the model. This design helps minimize randomness in the evaluation process. Then, for each question, our algorithm selects distractors based on the correct answer to generate $m$ options. In this way, we construct the final test samples and assemble AssoCiAm, which circumvents both internal and external ambiguity.



\subsection{AssoCiAm Analysis}
AssoCiAm consists of 2,025 test samples, each comprising an image with a corresponding Question-Options pair. The images are of high quality, with a resolution of 512, ensuring sufficient detail to effectively convey the necessary information. The benchmark includes three mT1 tasks with varying difficulty levels. AssoCiAm covers 25 classes, encompassing diverse aspects of daily life, ensuring its comprehensiveness. These key attributes make AssoCiAm a lightweight yet comprehensive benchmark for evaluating the associative ability of models. Table~\ref{tab:cir} provides an overview of the key statistics of AssoCiAm.
\begin{table}[htbp]
    \centering
    \begin{tabular}{|l|c|}
    \hline
    \textbf{Statistics}           & \textbf{Value} \\ \hline
    Class                      & 25 \\ 
    Image                      & 225 \\ 
    Image Resolution           & $512 \times 512$ \\ 
    4T1 Question           & 675 \\ 
    7T1 Question           & 675 \\ 
    10T1 Question          & 675 \\ \hline
    \end{tabular}
    \caption{Key statistics of AssCiAm. mTn question represents question with m options and n correct answers.}
    \label{tab:cir}
\end{table}


\section{Experiment Setup}
\label{sec:exp}
\subsection{Model Seclection}
\textbf{MLLMs.}
To comprehensively assess model performance, we consider a range of both open- and closed-source multimodal large language models. For each model family, we include a series of available models, ranging from good to better performance.
(i) InternVL2~\cite{chen2024internvl, chen2024far, gao2024mini, wang2024mpo, chen2024expanding},
(ii) LLaVA-Next~\cite{liu2023llava, liu2023improvedllava, liu2024llavanext, zhang2024llavanext-video, li2024llavanext-strong, li2024llavanext-ablations, li2024llava},
(iii) LLaVA-OneVision~\cite{li2024llava2},
(iv) MiniCPM-V~\cite{yao2024minicpm},
(v) Yi~\cite{ai2024yi},
(vi) CogVLM~\cite{wang2023cogvlm, hong2023cogagent, hong2024cogvlm2},
(vii) mPLUG-Owl~\cite{ye2023mplugowl, ye2023mplugowl2, ye2024mplugowl3longimagesequenceunderstanding},
(viii) Otter~\cite{li2023otter,li2023mimicit},
(ix) Qwen-VL~\cite{Qwen-VL},
(x) VisualGLM~\cite{ding2021cogview, du2022glm},
(xi) MiniGPT-4~\cite{zhu2023minigpt, chen2023minigptv2},
(xii) ChatGPT-4o~mini~\cite{hurst2024gpt},
(xiii) Gemini-1.5-pro~\cite{team2024gemini}.

\noindent{\textbf{Human experts.}}
To investigate human performance on this task, we invite three human experts to complete a quiz consisting of 10\% of our benchmark, randomly sampled from the benchmark.

\begin{figure*}[htbp]
  \includegraphics[width=0.2463\linewidth]{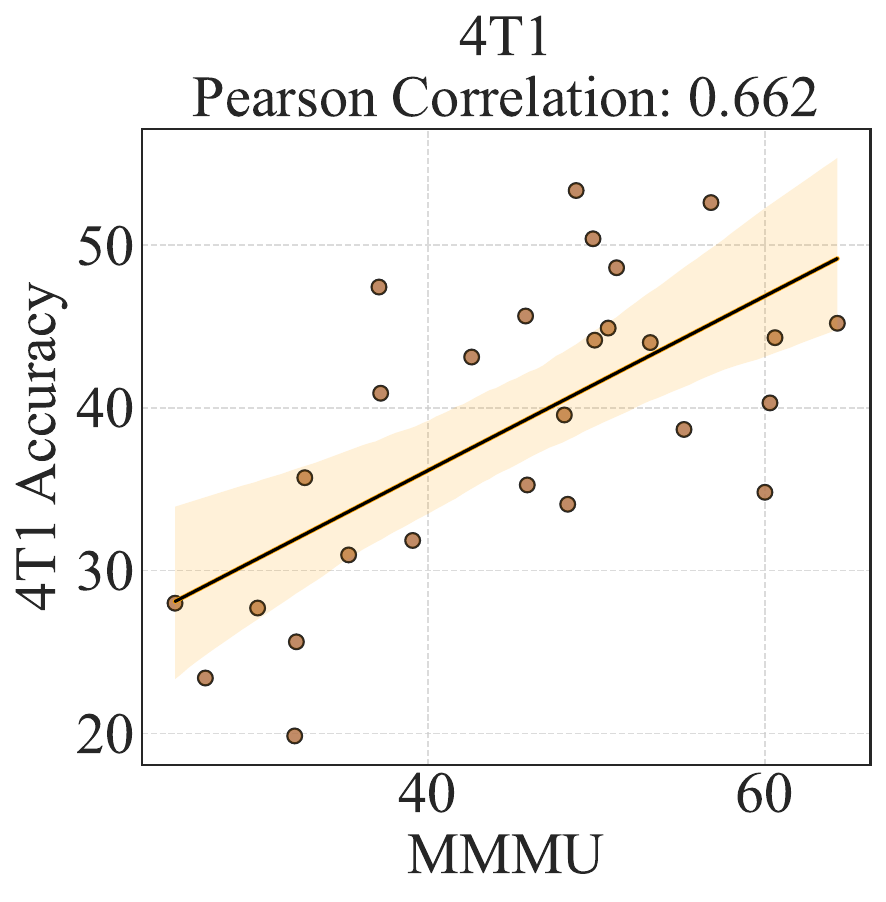} 
  \includegraphics[width=0.2463\linewidth]{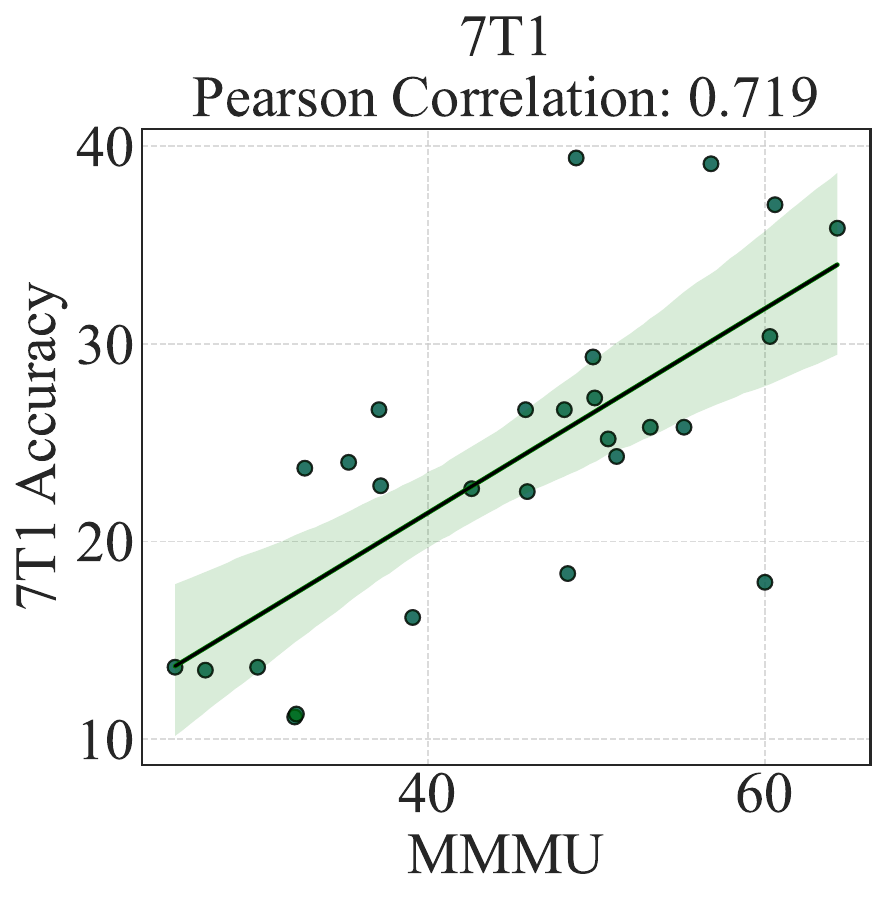} 
  \includegraphics[width=0.2463\linewidth]{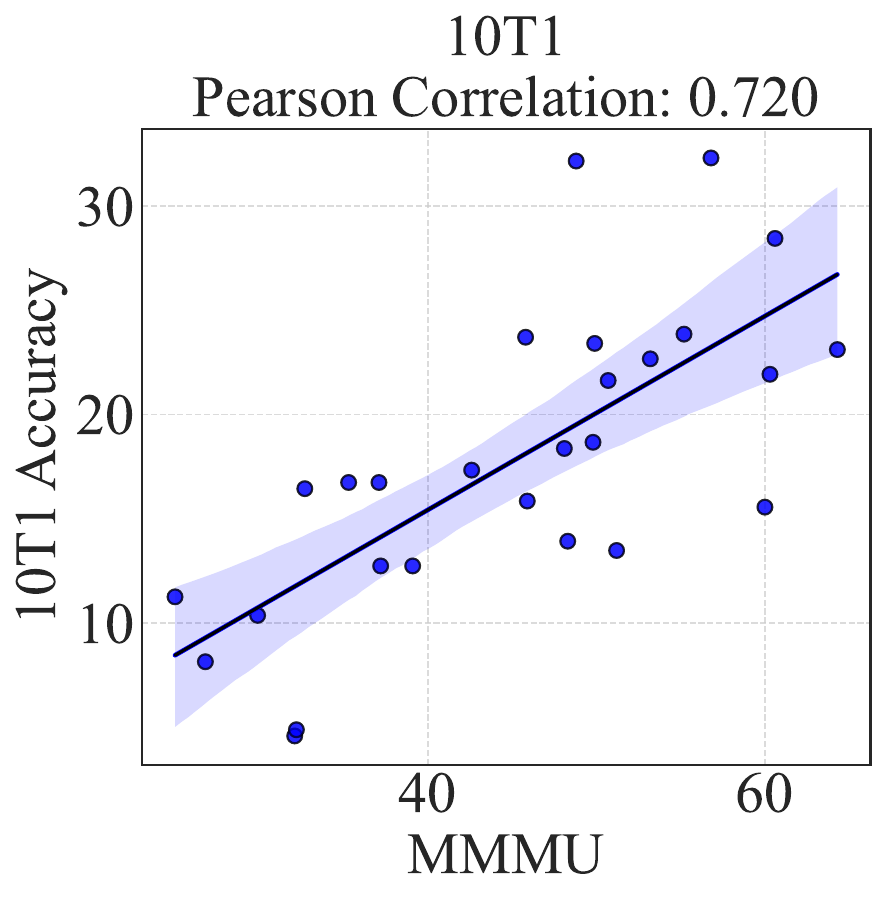} 
  \includegraphics[width=0.2463\linewidth]{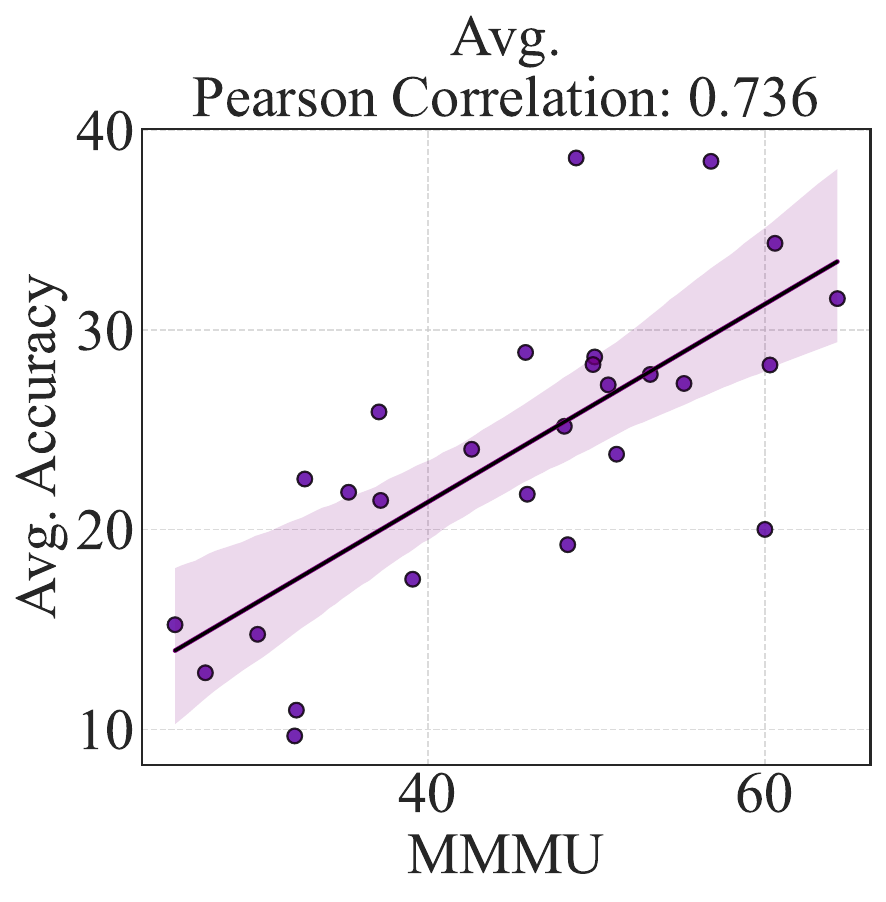}
  \caption {Linear regression lines illustrate the relationship between scores of association and cognition. mT1 represents association performance on mT1 tasks, while Avg. indicates the overall performance. The X-axis represents the cognition scores, and the Y-axis represents the association scores. The Pearson Correlation quantifies the correlation between association and cognition. The shaded area indicates the data points within a 95\% confidence interval around the regression line.}
    \label{fig:main}
\end{figure*}

\subsection{Evaluation Settings}
In our experiments, we use Top-1 accuracy as the evaluation metric, as it provides a fair measure for the multiple-option question-answering format and is widely adopted in prominent commonsense reasoning tasks~\cite{mihaylov2018can, talmor2018commonsenseqa, bisk2020piqa}. We evaluate model performance across three subtasks, reporting individual accuracy scores for each. To obtain a comprehensive understanding of models' associative abilities, we compute the weighted average accuracy across the three subtasks due to their different difficulty levels. The weight assigned to an mT1 task is $m$.

To ensure a fair evaluation, all models are tested using a unified prompt template within a two-shot demonstration setup. This setup is designed to stimulate associative reasoning across models.

\section{Results}
\label{sec:res}
\textbf{Overall Performance.}
\begin{table}[!t]
  \centering
  \resizebox*{0.99\linewidth}{!}{
    \begin{tabular}{lr|c|c|c|c}
    \toprule
    \textbf{Model} & \textbf{Size}  & \textbf{4T1}   & \textbf{7T1}   & \textbf{10T1}  & \textbf{Avg.} \\
    \midrule
    GPT-4o-mini & -     &  34.81     & 17.93      &  15.56     & 20.01 \\
    InternVL2-40B & 40B   & 38.67  & 25.78  & 23.85  & 27.32  \\
    InternVL2-26B & 26B   & 44.89  & 25.19  & 21.63  & 27.25  \\
    InternVL2-8B & 8B    & 48.59  & 24.30  & 13.48  & 23.77  \\
    InternVL2-4B & 4B    & 34.07  & 18.37  & 13.93  & 19.25  \\
    Qwen-VL-Max (0809) & -     & 45.19  & 35.85  & 23.11  & 31.56  \\
    Qwen-VL-Plus (0809) & -     & 44.00  & 25.78  & 22.67  & 27.77  \\
    LLaVA-OneVision-72B & 72B   & \underline{52.59}  & \underline{39.11}  & \textbf{32.30}  & \underline{38.43}  \\
    LLaVA-OneVision-7B & 7B    & \textbf{53.33}  & \textbf{39.41}  & \underline{32.15}  & \textbf{38.60}  \\
    LLaVA-Next-72B & 72B   & 44.15  & 27.26  & 23.41  & 28.64  \\
    LLaVA-Next-34B & 34B   & 39.56  & 26.67  & 18.37  & 25.17  \\
    LLaVA-Next-mistral-7B & 7B    & 30.96  & 24.00  & 16.74  & 21.87  \\
    MiniCPM-V 2.6 & 8B    & 50.37  & 29.33  & 18.67  & 28.26  \\
    MiniCPM-Llama3-V 2.5 & 8.5B  & 45.63  & 26.67  & 23.70  & 28.87  \\
    MiniCPM-V 2 & 2.8B  & 47.41  & 26.67  & 16.74  & 25.89  \\
    MiniCPM-V & 3B    & 40.89  & 22.81  & 12.74  & 21.46  \\
    Yi-vision & -     & 40.30  & 30.37  & 21.93  & 28.24  \\
    Yi-VL-34B & 34B   & 35.26  & 22.52  & 15.85  & 21.77  \\
    Yi-VL-6B & 6B    & 31.85  & 16.15  & 12.74  & 17.52  \\
    CogVLM-17B-chat & 17B   & 19.85  & 11.11  & 4.59  & 9.67  \\
    CogVLM2-19B-chat & 19B   & 43.11  & 22.67  & 17.33  & 24.02  \\
    VisualGLM-6B & 8B    & 27.70  & 13.63  & 10.37  & 14.76  \\
    MiniGPT-4-v2 & 8B    & 28.00  & 13.63  & 11.26  & 15.24  \\
    MiniGPT-4 (VicunaV0-13B) & 13B   & 23.41  & 13.48  & 8.15  & 12.83  \\
    mPLUG-Owl3 & 8B  & 34.22  & 23.70  & 19.26  & 23.59  \\
    mPLUG-Owl2 & 8.2B  & 35.70  & 23.70  & 16.44  & 22.53  \\
    OTTER-Image-MPT7B & 7B    & 25.63  & 11.26  & 4.89  & 10.96  \\
    Gemini-1.5-pro & -     & 44.30  & 37.04  & 28.44  & 34.33  \\
    \hline
    human & -     &   100.00    &  100.00     &  100.00     & 100.00  \\
    random & -     & 22.67  & 13.63  & 10.67  & 13.94  \\
    \bottomrule
    \end{tabular}%
    }
  \caption{The results of the evaluation. mT1 represents the score for a task with m options and one correct answer. The Avg. column represents the weighted average score, calculated by the formula: 
  $Avg. = \frac{(4 \times 4T1 + 7 \times 7T1 + 10 \times 10T1)}{4 + 7 + 10}$.}
  \label{tab:main}%
\end{table}%
The main results are presented in Table~\ref{tab:main}. LLaVA-OneVision 7B and LLaVA-OneVision 72B achieve the top-2 performances across the three tasks and on the overall average score, with only minor differences between them. However, all models do not achieve high scores and their scores decrease as the number of distractors increases.
These results underscore the challenges posed by the benchmark and indicate that mitigating ambiguity does not necessarily make the benchmark too easy for models. Moreover, models still exhibit a significant gap compared to human performance. This behavior highlights a discrepancy between models and humans in associative tasks. Notably, human participants consistently demonstrate robust associative abilities even as task difficulty increases, whereas model performance declines under similar conditions. The inconsistency observed across model responses indicates that models lack a strong ability to form meaningful connections between seemingly unrelated concepts—the core of association. This limitation further underscores the gap between current models and human-level associative capabilities.

\noindent{\textbf{Correlation between cognition and association.}}
Previous studies~\cite{martinsen1994effect, martinsen1993insight, kaufmann1979explorer, runco1995cognition, huang2025causality2, mednick1962associative} have highlighted that human cognitive abilities (knowledge, perception, and reasoning) form the foundation of associative ability (e.g., association requires adequate knowledge, the ability to perceive each concept clearly, and the capacity to reason from one concept to another.). These theories suggest that for both humans and models, association and cognition should exhibit at least some degree of correlation. 

To analyze the correlation between cognition and association, we use scores from the MMMU benchmark~\cite{yue2024mmmu} as an indicator of models' cognitive ability, as MMMU is specifically designed to evaluate comprehensive cognitive skills of MLLMs. The Pearson Correlation Coefficient~\cite{sedgwick2012pearson,schober2018correlation,cohen2009pearson} is computed to quantify the linear correlation between cognition and association. 
As shown in Fig.~\ref{fig:main}, all Pearson Correlation Coefficient values exceed 0.66, with three exceed 0.71, demonstrating a strong positive correlation between cognition and association. Most data points fall within the confidence interval, further supporting the linear relationship. Consistent with the perspective, the results show a strong correlation between association and cognition, aligning well with human cognitive theories. This further validates that our benchmark is well designed in accordance with human cognitive principles.


\begin{figure}[htbp]
  \includegraphics[width=1\linewidth]{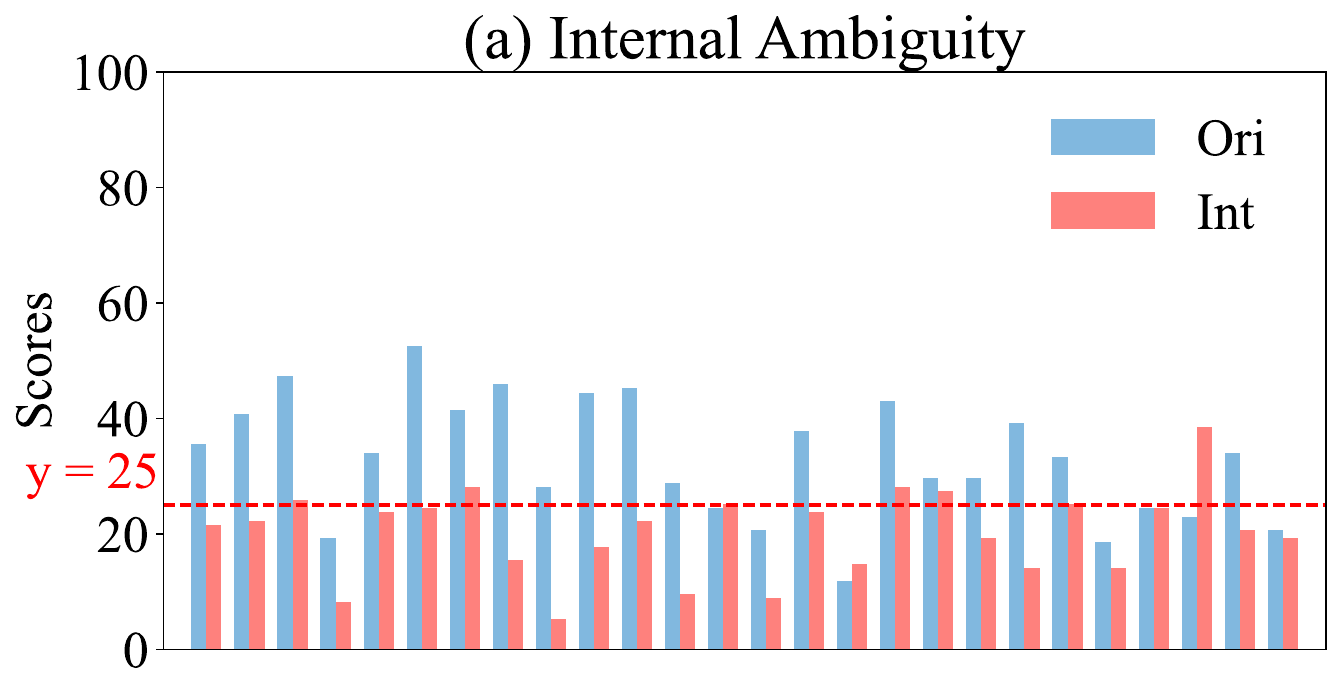}
  \includegraphics[width=1\linewidth]{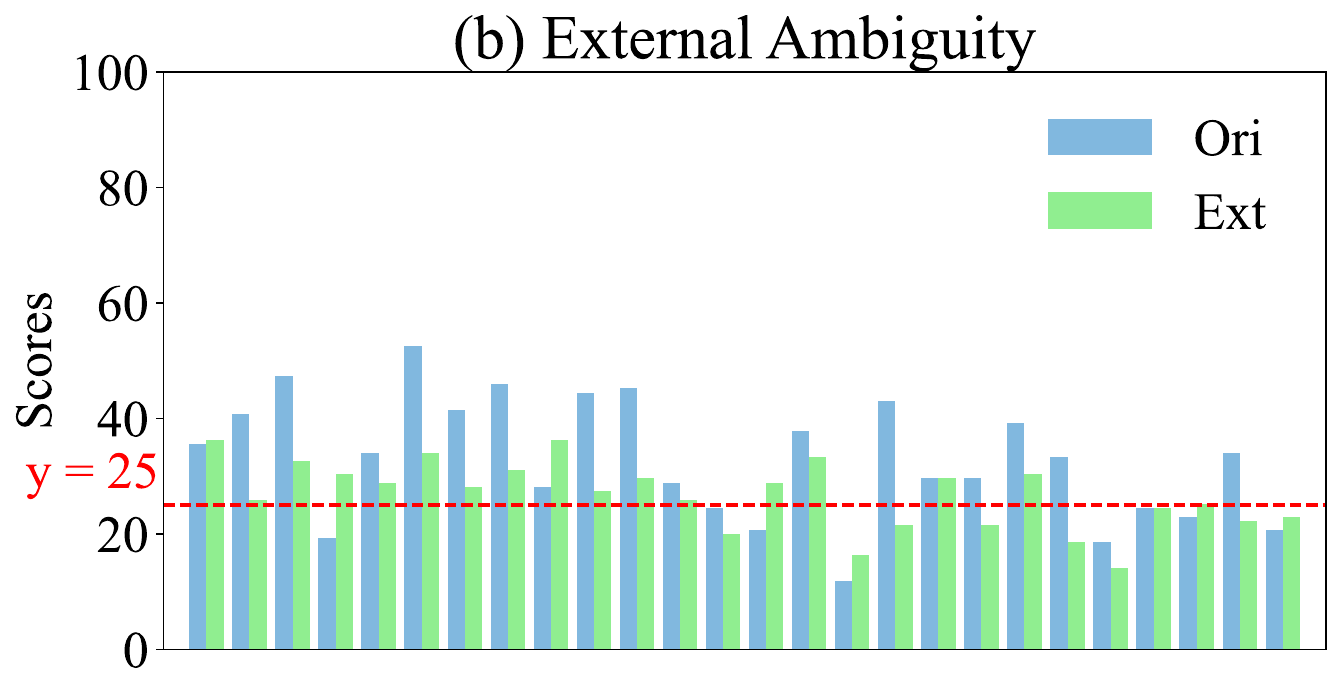}
  \caption {The evaluation results of models. Ori indicates the scores evaluated on Ori, Int indicates the scores evaluated on Int and Ext indicates the scores evaluated on Ext. The red line $y=25$ indicates the expected score of answering questions randomly.}
    \label{fig:int}
\end{figure}

\section{Analysis}
\label{sec:ana}
To analyze the impact of ambiguity and the effectiveness of our method, we focus on the following four research questions (RQs):
\begin{itemize}[leftmargin=0.3cm]
\vspace{-0.3cm}
    \item RQ1: How does the presence of internal ambiguity affect the evaluation outcomes?
\vspace{-0.3cm}
    \item RQ2: How does the presence of external ambiguity affect the evaluation outcomes?
\vspace{-0.3cm}
    \item RQ3: Does DINO-v2~\cite{oquab2023dinov2, darcet2023vitneedreg} focus on geometric shapes of masks?
    
    
\vspace{-0.3cm} 
    \item RQ4: Does our algorithm effectively mitigate external ambiguity in distractor selection process?  
\end{itemize}

\noindent{\textbf{The ambiguity affects association evaluation~(RQ1\&RQ2).}}
To investigate the impact of ambiguity, we first sample classes and their associative questions with four options to construct an original validation set~(Ori), a subset of AssoCiAm. 
Subsequently, we derive two validation sets from Ori: one with \textbf{internal ambiguity}~(Int) and one with \textbf{external ambiguity}~(Ext).

\begin{figure}[htbp]
    \includegraphics[width=0.157\textwidth]{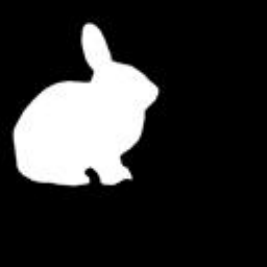} 
    \includegraphics[width=0.157\textwidth]{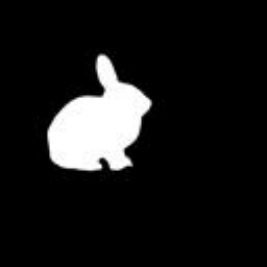}
    \includegraphics[width=0.157\textwidth]{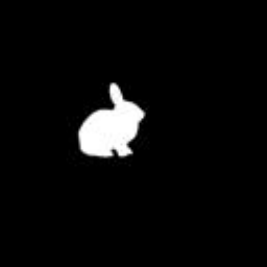} 
    \caption{Examples of a series of scale masks.}
    \label{fig:sc_pic}
\end{figure}

\begin{figure*}[htbp]
  \includegraphics[width=0.33\linewidth]{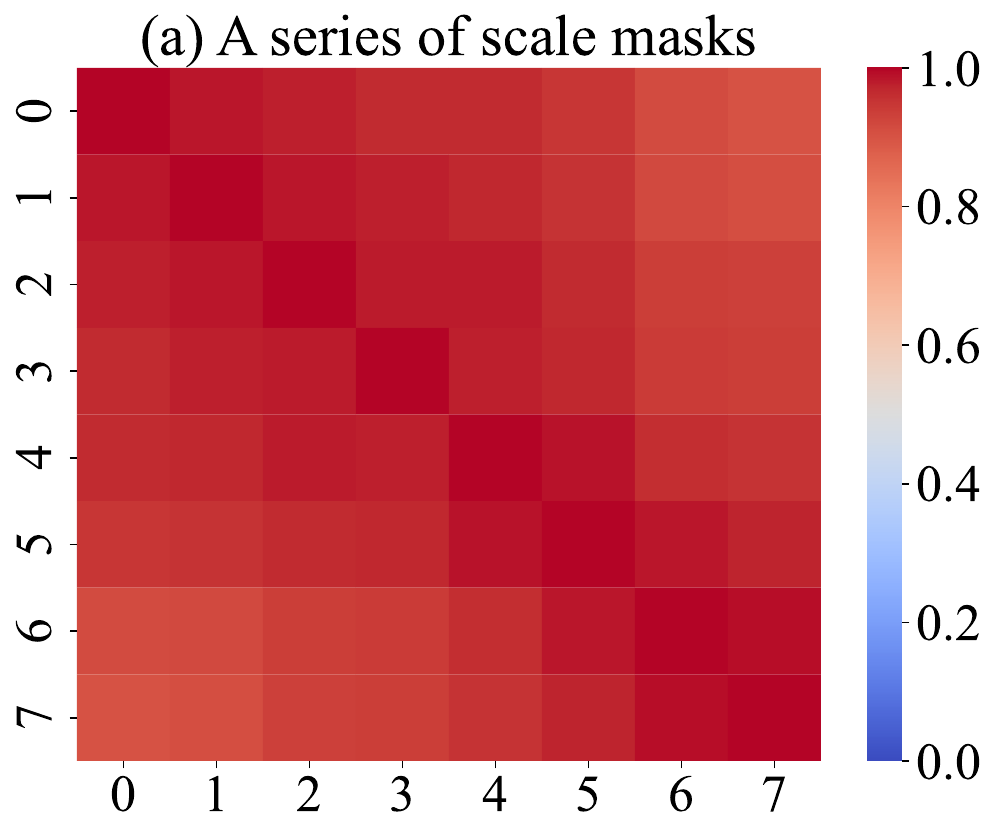}
  \includegraphics[width=0.33\linewidth]{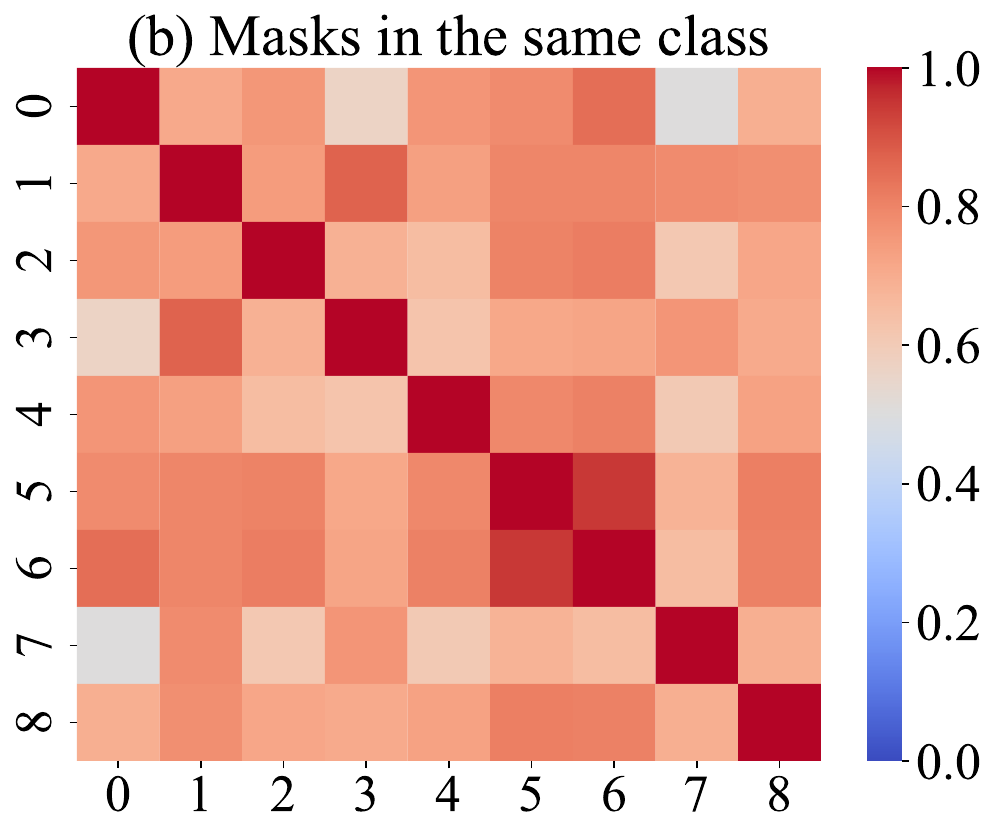}
  \includegraphics[width=0.33\linewidth]{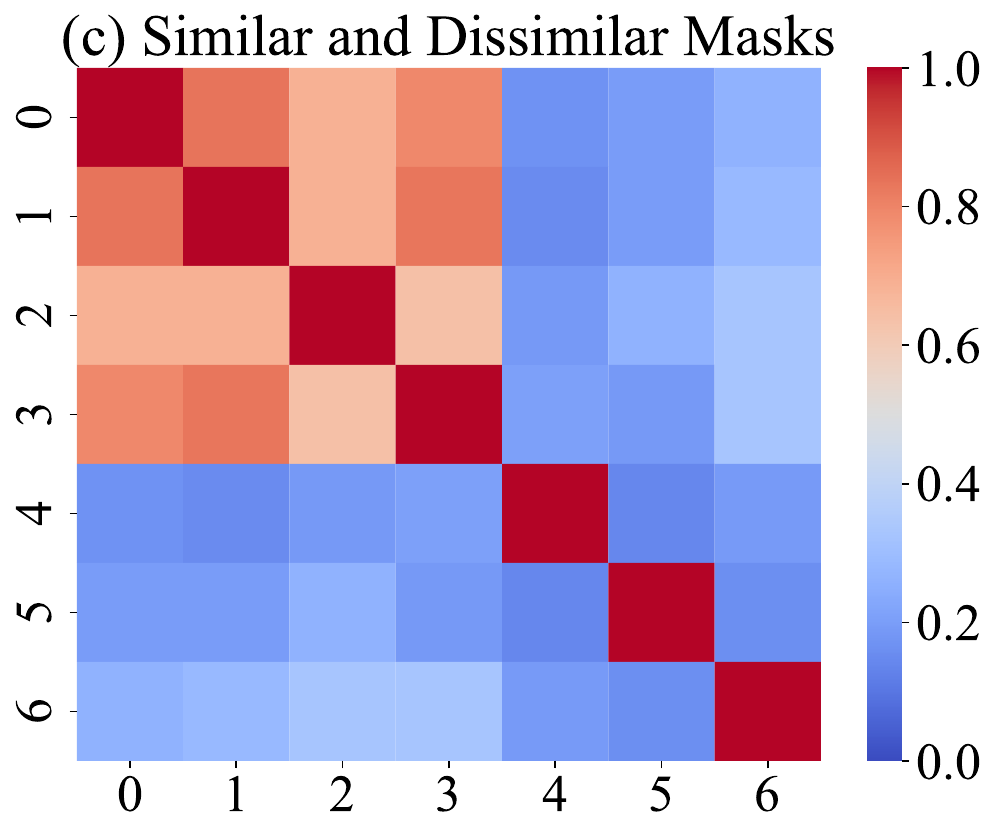}
  \caption {Heatmaps of similarity among masks. A similarity score of 1 indicates a high degree of similarity, while a score of 0 indicates a high degree of dissimilarity. The number in axis denotes the id of masks. (a) illustrates the similarity among a series of scale rabbit masks. (b) illustrates similarity among different chrysanthemum masks. In (c), id~3 is the shark mask; id~0~to~2 are masks of  ambiguous distractors; id~4~to~6 are masks of unambiguous distractors. See Appendix~\ref{app:dino} for more results.}
    \label{fig:hm}
\end{figure*}

For Int, we invite human experts to replace the original answers in Ori with unreasonable answers and ensure that options remain dissimilar, to explicitly introduce internal ambiguity. 
For instance, the original answer, 'A.~Goblet,' is replaced with 'A.~Blackboard,' and it still serves as the correct answer.
Due to external ambiguity avoidance in the Ori, this process creates Int, where internal ambiguity is the primary source of ambiguity.

To study the effects of internal ambiguity comprehensively, we assess performance of models in Section~\ref{sec:res} on Ori and Int. As shown in Fig.~\ref{fig:int}~(a), most of the scores in Int tend to approximate random-choice performance levels, compared to those in Ori. This finding suggests that internal ambiguity renders models incapable of distinguishing answers based on questions. Additionally, the result of scores nearing random performance highlights the lack of faithfulness in these evaluations, as models fail to answer questions through meaningful associations and answer questions randomly.

For Ext, we invite human experts to add new classes to explicitly construct external ambiguity. 
To achieve this, the human experts follow this procedure: observe the masks in the \textbf{original mask set}~(Section~\ref{sec:filter}), use association to infer what the masks resemble, and generate ambiguous distractors based on these associations. By replacing the distractors in Ori with these ambiguous distractors, we construct Ext. In this set, all distractors are ambiguous, while all answers remain reasonable, making external ambiguity the primary risk.

As shown in Fig.~\ref{fig:int}~(b), we use the same models as in RQ1 and observe that most of the scores in Ext also approach random-choice levels compared to Ori. This result indicates that models are confused by the choices due to all options being plausible answers and fail to answer questions correctly, even though they make meaningful associations.

\noindent{\textbf{DINO-v2 calculate the similarity among masks~(RQ3).}}
To assess whether DINO-v2 primarily relies on geometric shapes, we scale masks to generate a series of versions, as shown in Fig.~\ref{fig:sc_pic}, and calculate their pairwise similarity.
The result, presented in Fig.~\ref{fig:hm}~(a), shows high similarity values, indicating that DINO-v2 primarily focuses on geometric shapes.


Next, we evaluate whether DINO-v2 produces consistent similarity representations for masks within the same semantic class, despite minor geometric variations. Human experts are invited to select representative masks from filtered set~(Section~\ref{sec:filter}), following the \textbf{key principle} outlined in Section~\ref{sec:filter}. DINO-v2 is then employed to calculate the similarity among these masks belonging to the same class. As shown in Fig.~\ref{fig:hm}~(b), the similarity among these representative masks is high and exceeds threshold 50\%, indicating that DINO-v2 exhibits robustness to small shape differences within the same class.


Finally, we examine whether DINO-v2 can capture shape-level similarity across different semantic classes.
To this end, human experts gather representative masks of ambiguous distractors from RQ2 to create an \textbf{extension set}. We sample classes and calculate similarity among their own masks, the masks of their ambiguous distractors, and the masks of their unambiguous distractors. The results, illustrated in Fig.~\ref{fig:hm}~(c), show that similarity among shark and its ambiguous distractors is high, whereas the similarity scores between shark and its unambiguous distractors are much lower. Furthermore, the similarity between shark and its ambiguous distractors is significantly higher than that between shark and its unambiguous distractors. These results indicate that DINO-v2 effectively  measures geometric similarity between masks from both similar and dissimilar classes.

Overall, our findings demonstrate that DINO-v2 predominantly encodes geometric shape information, enabling robust assessment of visual similarity across a wide range of mask classes.

\noindent{\textbf{The algorithm circumvents external ambiguity~(RQ4).}}
In Section~\ref{sec:filter}, we invited human experts to refine the masks and images to ensure the avoidance of internal ambiguity. So, we only analyze the effectiveness of the algorithm in Section~\ref{sec:alg}. In this experiment, we use the extension set~(RQ3) and manually label ambiguous distractors for each class in the original mask set~(Section~\ref{sec:filter}). For each class in the original set, we perform two processes: randomly selecting distractors from the extension set, repeated ten times, and applying the algorithm to select distractors, also repeated ten times. An answer and its corresponding distractors form a multiple-option set. We then count the number of these multiple-option sets that contain ambiguous distractors. As shown in Table~\ref{tab:cir2}, 15\% of the randomly selected multiple-option sets overall include ambiguous distractors. For each option quantity, the proportion of randomly selected multiple-option sets containing ambiguous distractors increases as the number of options grows. In contrast, none of the algorithm-selected sets contain any ambiguous distractors. These results clearly highlight the effectiveness of the algorithm in avoiding external ambiguity.

\begin{table}[htbp]
  \centering
    \begin{tabular}{c|c|c}
    \toprule
    \textbf{Quantity} & \textbf{Random} & \textbf{Algorithm} \\
    \midrule
    4     & 9.60\% & 0.00\% \\
    7     & 17.20\% & 0.00\% \\
    10    & 18.00\% & 0.00\% \\
    overall & 15.00\% & 0.00\% \\
    \bottomrule
    \end{tabular}%
  \caption{Comparison of distractor selection methods: random selection and algorithmic selection. Quantity refers to the number of options in a multiple-option set. Random represents the overall proportion of multiple-option sets with ambiguous distractors when selected randomly, while Algorithm represents the proportion when selected via the algorithm.}
  \label{tab:cir2}%
\end{table}%

\section{Conclusion}
\label{sec:conclu}
This paper identifies the inherent answer ambiguity in association evaluation and validates its significant and unavoidable impact on evaluation reliability. To address this issue, we propose AssoCiAm, a benchmark constructed through a hybrid computational method designed to mitigate ambiguity. Our evaluation of MLLMs on AssoCiAm reveals that these models still exhibit a noticeable gap compared to human performance in associative tasks. Furthermore, we find that associative ability is closely related to cognitive capability, highlighting the importance of advancing both aspects in future MLLM development.

\section*{Limitations}
\label{sec:limit}
While our tasks focus on association based on object shapes to reflect this ability from a specific perspective, we aim to expand association evaluation into a more comprehensive and broader framework in the future. 
Additionally, during the evaluation process, differences in training methods and data among MLLMs may result in varied interpretations of images and options. This could lead to models selecting the correct answer by excluding distractors through understanding the similar semantics among the distractors rather than directly solving the task as intended. 
Finally, our benchmark assigns a single correct answer to each question. However, considering that ambiguity arises from answer diversity, future work should explore more evaluation frameworks, such as multiple-correct-option or open-ended questions, which can not only mitigate the answer ambiguity but also preserve and leverage answer diversity serving as an intrinsic property of association evaluation.

\section*{Acknowledgements}
This work is supported by National Natural Science Foundation of China under Grants No. 623B2099, No. 62206314 and Science and Technology Projects in Guangzhou under Grant No. 2024A04J4388.

\bibliography{custom}

\appendix
\clearpage
\section{Appendix}
\label{sec:appendix}
\subsection{Details for collecting images}
To collect images, whose classes are applicable to the associative task, we first merge similar classes into a single category to address subtle differences between certain classes, such as cocks and hens. Additionally, classes that are uncommon for human recognition or lack distinct shapes are excluded. 

\subsection{Details for extracting masks}
\label{app:mask}
\begin{figure}[htbp]
  \includegraphics[width=\linewidth]{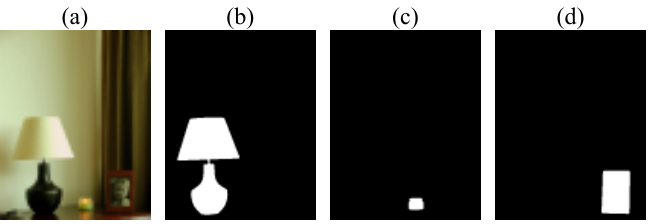} 
  \caption {An example of masks extracted by SAM~\cite{kirillov2023segany}. (a) represents the original image of a table lamp. (b) shows the mask corresponding to the table lamp, which is the accurately segmented mask. (c) and (d) illustrate redundant masks extracted from other objects in the image (a), namely, a hand cream and a photo frame.}
    \label{fig:redun}
\vspace{-0.2cm}
\end{figure}

As the segmentation process operates in batches, SAM~\cite{kirillov2023segany} generates multiple masks per image, which may include both class-relevant masks and redundant masks. For instance, Fig.~\ref{fig:redun}~(b) illustrates a representative mask from Fig.~\ref{fig:redun}~(a), while Fig.~\ref{fig:redun}~(c, d) shows redundant masks.

\begin{figure}[htbp]
\vspace{-0.3cm}
  \includegraphics[width=\linewidth]{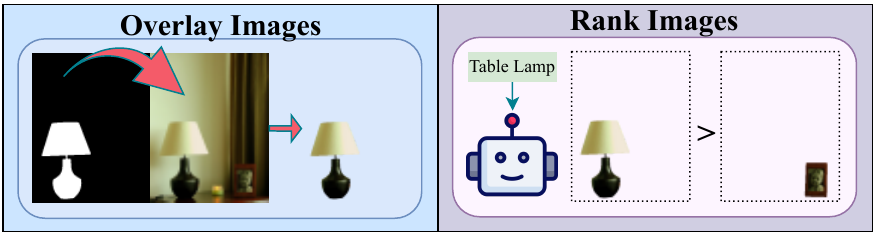} 
  \caption {An overview of eliminating redundant masks}
    \label{fig:rank}
\vspace{-0.5cm}
\end{figure}

To eliminate redundant masks, the following approach is adopted: for each image, model CLIP~\cite{radford2021learning} is utilized to rank a series of images from overlaying segmented masks onto the original image, as illustrated in Fig.~\ref{fig:rank}. The mask corresponding to the top-ranked image is then selected as the class-relevant one. This approach ensures the accurate extraction of masks per image.

\begin{figure}[htbp]
\vspace{-0.3cm}
  \includegraphics[width=0.49\linewidth]{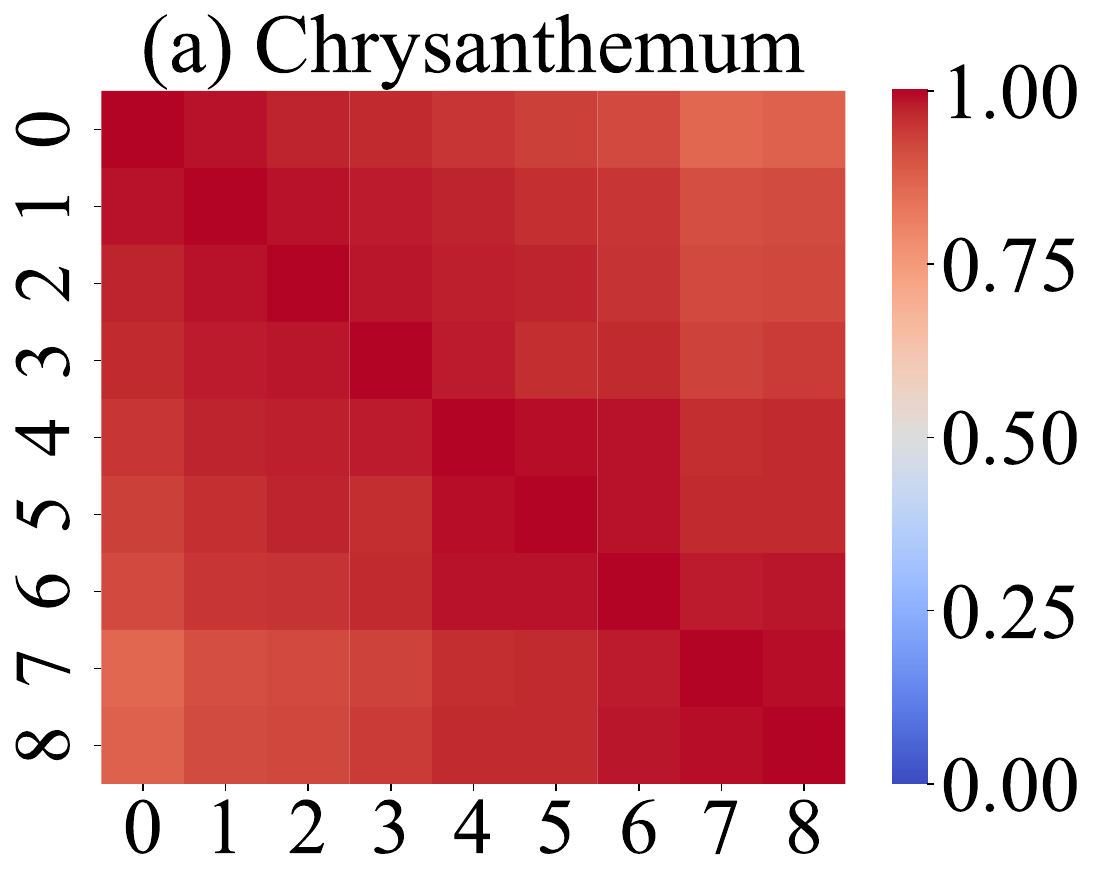} 
  \includegraphics[width=0.49\linewidth]{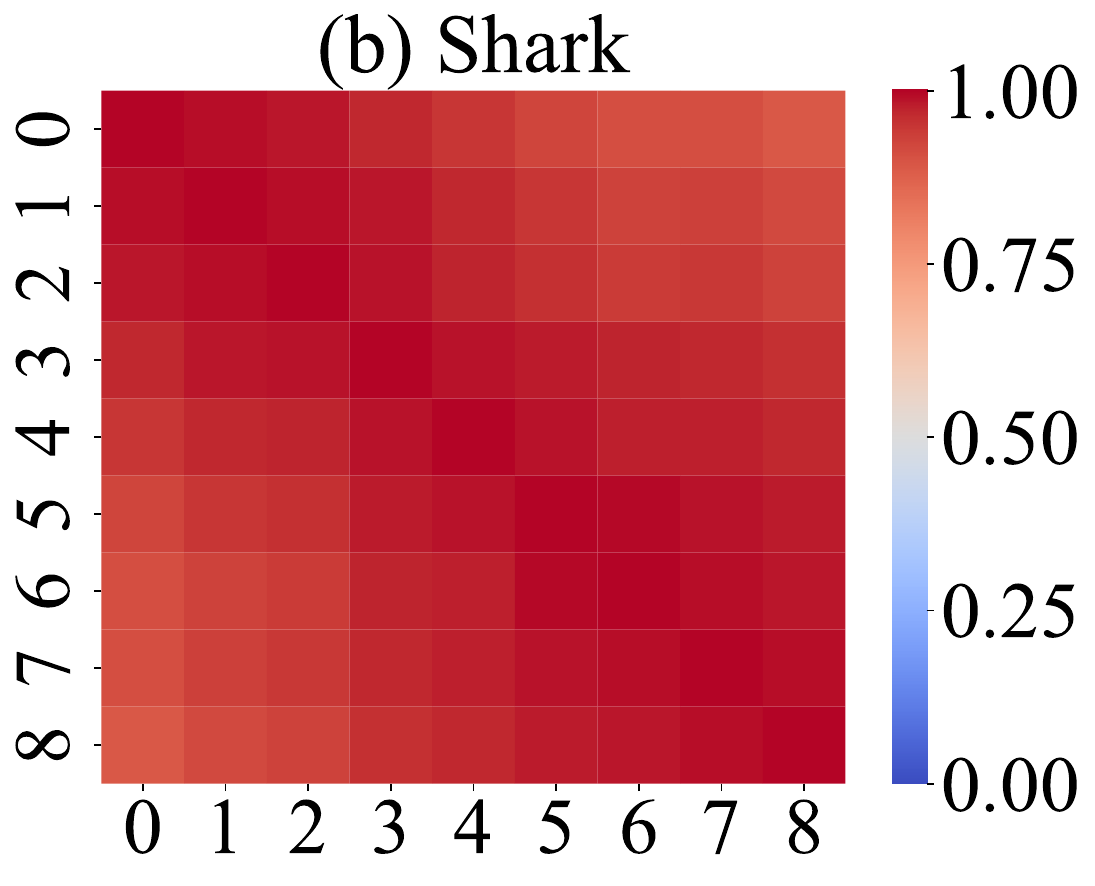} 
  \includegraphics[width=0.49\linewidth]{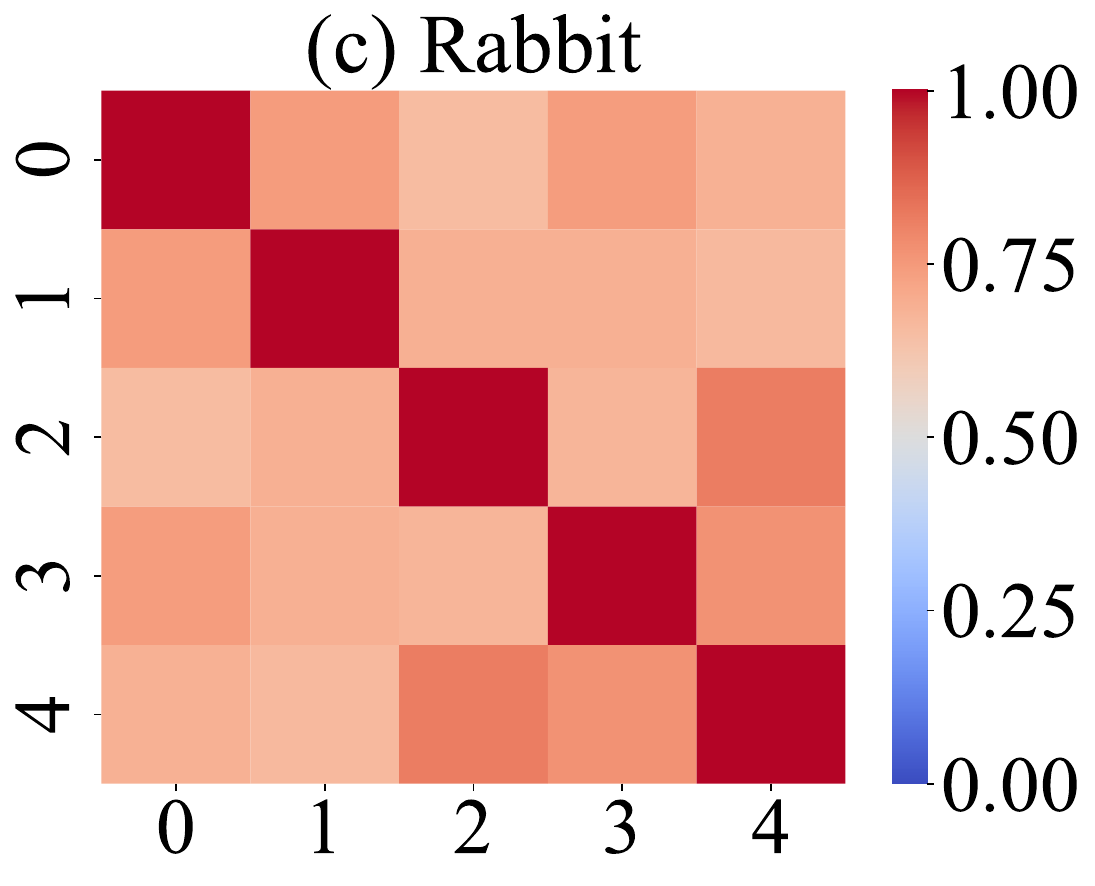} 
  \includegraphics[width=0.49\linewidth]{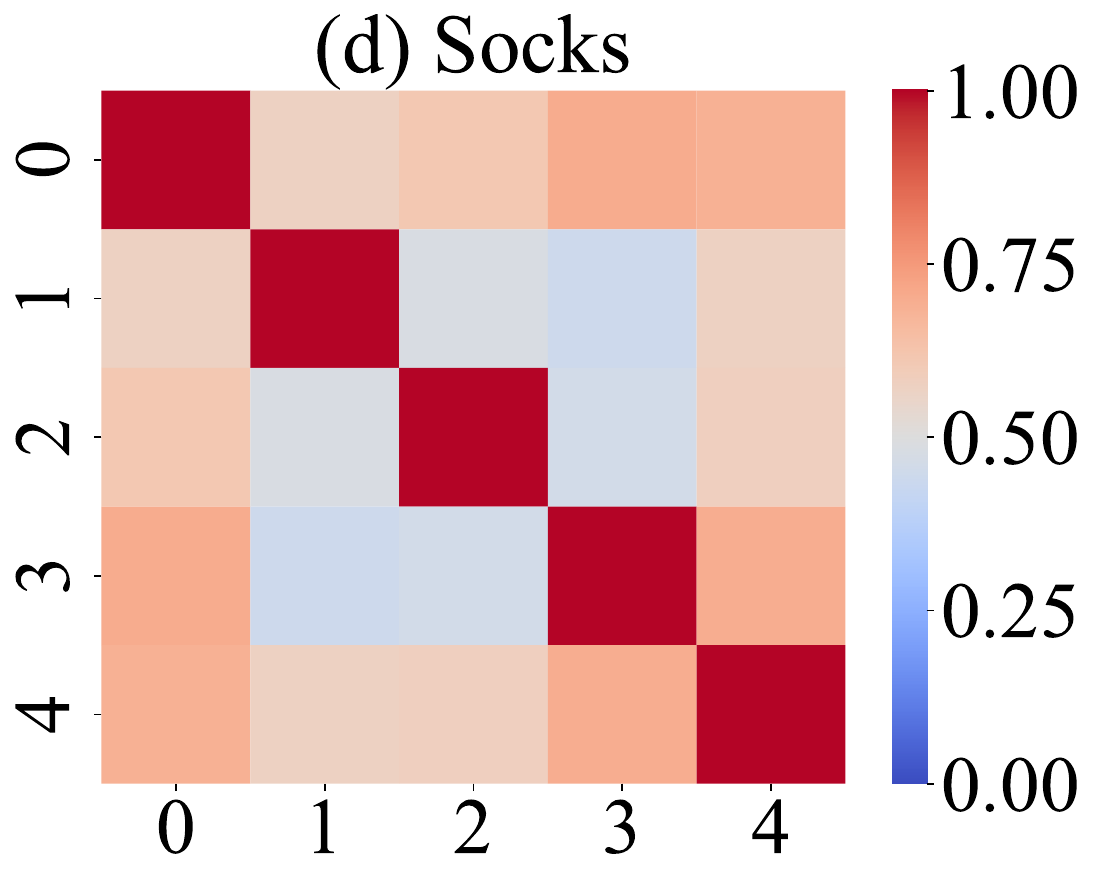} 
  \includegraphics[width=0.49\linewidth]{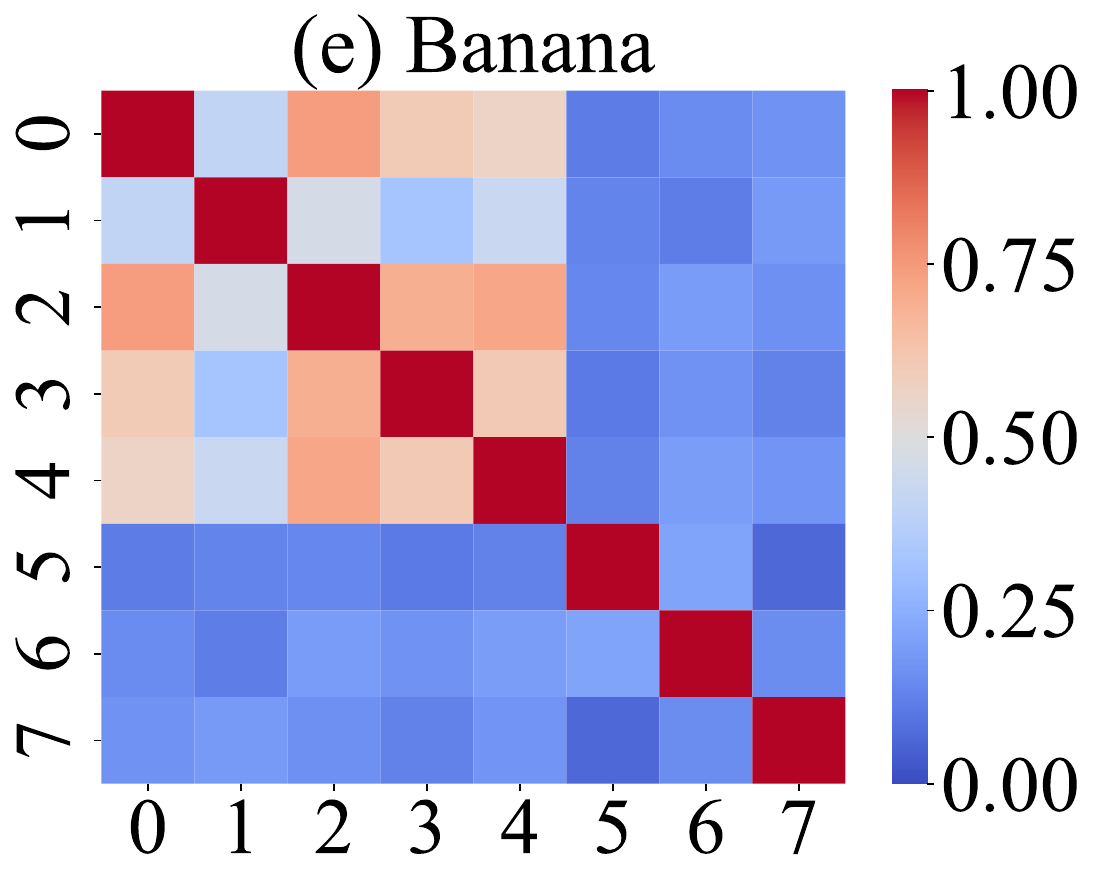} 
  \includegraphics[width=0.49\linewidth]{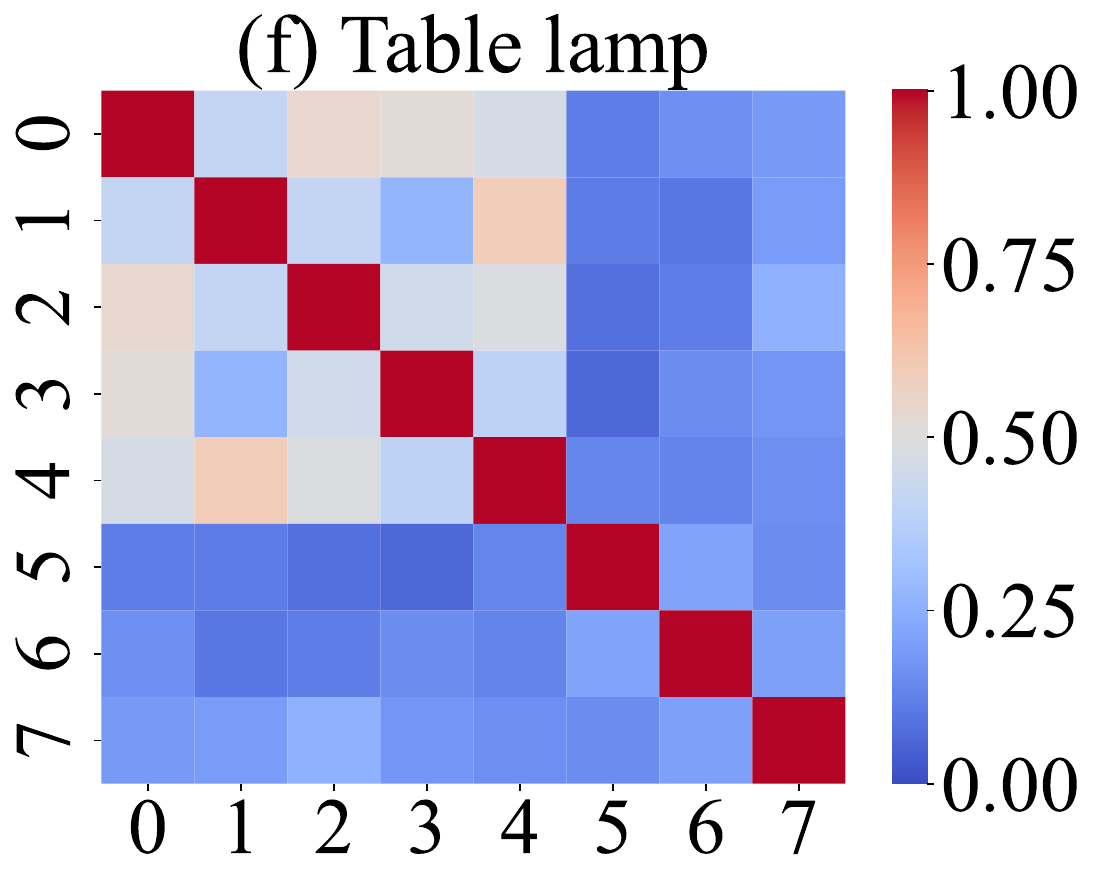} 
  \caption {(a, b) are similarities among a series of scale masks. The masks belong to chrysanthemum or shark. (c, d) are similarities among masks belonging to the same class. In (e, f), id 0,1,2,3 indicates masks of ambiguous distractors, id 5,6,7 are masks unambiguous distractors. id 4 is the masks of correct answer.}
    \label{fig:dinov2}
\vspace{-0.2cm}
\end{figure}


\subsection{Details for the Genetic Algorithm}
\label{app:ga}
The optimization of the function~\ref{eq:final} is an NP-hard problem. To address this, we adopt a genetic algorithm, a classic randomized search method, that is widely used for solving NP-hard problems. Below are the main details of our implementation:

\textbf{Encoding and Initialization.}
Each solution is represented as a binary string, where 1 indicates a question is selected and 0 indicates it is not. We initialize the population with 50 randomly generated selection combinations.

\textbf{Parent Selection.}
We use a tournament selection strategy to choose parent pairs based on fitness scores, which is calculated by function~\ref{eq:final}.

\textbf{Searching.}
For each parent pair, we apply single-point crossover with a probability of 0.8, i.e.,  $$\text{offspring} = \text{parent1}[ : \text{point} ] + \text{parent2}[ \text{point} : ].$$
With a probability of 0.2, the offspring is directly copied from one of the parents without crossover. Each offspring undergoes bit-flip mutation with a probability of 0.05. To satisfy the selection constraints, we randomly flip excess 0s to 1s or vice versa in the offspring if needed. This process is repeated for a fixed number of generations to optimize the objective function~\ref{eq:final}.

\subsection{Algorithm in Section~\ref{sec:alg}}
We summarize our algorithm and the process of our algorithm is shown in Alg.~\ref{alg:mini}
\begin{algorithm}[th]  
    \caption{External Ambiguity Avoidance}
    \raggedright
    \textbf{Input:}  The masks set $V$ and answer mask $v_0$. The regularization parameter $\lambda$. The number of masks $n$ and choices $m$. 

    \textbf{Output:} The optimal set of distractors and answers $V'^*$
    \begin{algorithmic}[1]
    \State Initialize graph $G$: $e_{ij}$ = DINO-v2 ($v_i$, $v_j$) 
    \State Construct objective function $F(G')$ in Eq.~(\ref{eq:final})
    \State Optimal $F(G')$ via GA$(n, m, G, v_0, \lambda)$  
    
    \algorithmiccomment{GA: Genetic Algorithm} 
    \State Select the $G'^*$ = argmin$_{G'}F(G')$
    \State Decode $V'^*$ from $G'^*=<V'^*, E'^*>$
    \State\Return $V'^*$ 
    \end{algorithmic} 
    \label{alg:mini}   
\end{algorithm}

\subsection{DINO-v2 calculate the similarity}
\label{app:dino}
As shown in Fig.~\ref{fig:dinov2}~(a, b), the similarities among a series of scale masks are consistently high. Additionally, Fig.~\ref{fig:dinov2}~(c, d) demonstrates that similarities among masks within the same class are predominantly above 70\%, with the lowest values still close to 50\%. Fig.~\ref{fig:dinov2}~(e, f) highlights a clear distinction between similar and dissimilar masks: similar masks exhibit higher similarity scores, while dissimilar masks have notably lower scores. Although in Fig.~\ref{fig:dinov2}~(f), the similarities between the correct answer masks and ambiguous distractor masks are not as high as in other cases, they remain significantly higher than those between the correct answer masks and unambiguous distractor masks. These results underscore that DINO-v2 primarily focuses on the geometric shapes of masks.


\subsection{$\lambda$ balances $S(G')$ and $\sigma^2(G')$}
We analyze the effect of $\lambda$ using the original mask set, as selecting an appropriate $\lambda$ for constructing AssoCiAm depends on the properties of the original set. A series of $\lambda$ values are tested by sampling some masks as $v_0$ and applying our algorithm to calculate the corresponding $S(G')$ and $\sigma^2(G')$. As shown in Table~\ref{tab:trade}, increasing $\lambda$ results in an upward trend in $S(G')$ and a downward trend in $\sigma^2(G')$. With a well-chosen $\lambda$, $S(G')$ increases moderately, while $\sigma^2(G')$ decreases significantly, achieving a balanced trade-off between the two metrics.

\begin{table}[t]
  \centering
  \resizebox*{0.999\linewidth}{!}{
    \begin{tabular}{c|c|c|c|c|c}
    \toprule
    V$_0$    & $\lambda$ & $S(G')$   & $\sigma^2(G')$   & $\Delta S(G')$ & $\Delta\sigma^2(G')$ \\
    \midrule
    \multirow{5}[2]{*}{12} & 0     & \textbf{0.1278} & 0.0062 & 0.00\% & 0.00\% \\
          & 0.5   & \textbf{0.1278} & 0.0062 & 0.00\% & 0.00\% \\
          & 1     & 0.1309 & 0.0029 & 2.39\%$\uparrow$ & -52.72\%$\downarrow$ \\
          & 2     & 0.1309 & 0.0029 & 2.39\%$\uparrow$ & -52.72\%$\downarrow$ \\
          & 5     & 0.1350 & \textbf{0.0012} & 5.62\%$\uparrow$ & -79.83\%$\downarrow$ \\
    \midrule
    \multirow{5}[2]{*}{17} & 0     & \textbf{0.1423} & 0.0087 & 0.00\% & 0.00\% \\
          & 0.5   & \textbf{0.1423} & 0.0087 & 0.00\% & 0.00\% \\
          & 1     & 0.1468 & 0.0034 & 3.10\%$\uparrow$ & -60.31\%$\downarrow$ \\
          & 2     & 0.1468 & 0.0034 & 3.10\%$\uparrow$ & -60.31\%$\downarrow$ \\
          & 5     & 0.1557 & \textbf{0.0010} & 9.42\%$\uparrow$ & -89.02\%$\downarrow$ \\
    \midrule
    \multirow{5}[2]{*}{18} & 0     & 0.\textbf{1503} & 0.0081 & 0.00\% & 0.00\% \\
          & 0.5   & 0.1510 & 0.0032 & 0.45\%$\uparrow$ & -60.63\%$\downarrow$ \\
          & 1     & 0.1510 & 0.0032 & 0.45\%$\uparrow$ & -60.63\%$\downarrow$ \\
          & 2     & 0.1535 & \textbf{0.0011} & 2.17\%$\uparrow$ & -86.44\%$\downarrow$ \\
          & 5     & 0.1535 & \textbf{0.0011} & 2.17\%$\uparrow$ & -86.44\%$\downarrow$ \\
    \midrule
    \multirow{5}[2]{*}{25} & 0     & \textbf{0.1601} & 0.0233 & 0.00\% & 0.00\% \\
          & 0.5   & 0.1610 & 0.0036 & 0.53\%$\uparrow$ & -84.56\%$\downarrow$ \\
          & 1     & 0.1610 & 0.0036 & 0.53\%$\uparrow$ & -84.56\%$\downarrow$ \\
          & 2     & 0.1610 & 0.0036 & 0.53\%$\uparrow$ & -84.56\%$\downarrow$ \\
          & 5     & 0.1712 & \textbf{0.0009} & 6.93\%$\uparrow$ & -96.27\%$\downarrow$ \\
    \bottomrule
    \end{tabular}%
    }
  \caption{The values with different $\lambda$. $V_0$ indicates the id of the answer. $\Delta S(G')$ and $\Delta \sigma^2(G')$ are the the change of $S(G')$ or $\sigma^2(G')$ relative to that calculated with $\lambda = 0$. $\Delta S(G')= \frac{(S(G') - S(G')_{\lambda = 0})}{S(G')_{\lambda = 0})}$; $\Delta \sigma^2(G')= \frac{(\sigma^2(G') - \sigma^2(G')_{\lambda = 0})}{\sigma^2(G')_{\lambda = 0})}$ }
  \label{tab:trade}%
\end{table}%

\subsection{Prompt}
Prompt template used in evaluation is shown below:\\
\textit{You are a test subject participating in an associative ability test. Your task is as follows: Given an image and a question, you need to fully utilize your associative abilities and choose the best option from the given choices to answer the question. The given image and multiple-choice question are in the INPUT, specifically:}\\
\textit{1.IMAGE: A provided image}\\
\textit{2.QUESTION: A given question related to the image, requiring association to answer}\\
\textit{3.OPTION: The given options, from which you need to select the best option based on your associative thinking to answer the question}\\
\textit{Your OUTPUT should include ANSWER, formatted as follows:}\\
\textit{A/B/C/D}\\
\textit{One of the letters A, B, C, or D, representing the best option for answering the question.}\\
\textit{Please strictly follow this format to output the answer for the question posed in the INPUT}\\
\\
\textit{Associative thinking helps people understand things from different perspectives, enabling them to quickly adapt to new situations and improve problem-solving abilities. In the field of LLM, associative ability is also a very important metric. Now, we will test your associative thinking. Based on the given image, fully utilize your associative abilities to think about the given question, and select the best option from the given choices as your answer. Then, output your answer in the specified format.}\\
\\
\textit{Below are some examples. Please strictly follow the format.}\\
\\
\textit{Example 1:}\\
\textit{INPUT: \{}

	\textit{"IMAGE": ,}
    
	\textit{"QUESTION": "Observe the mountain in the image and use your associative thinking to consider what it resembles. Please choose the best option from the following choices to answer the question.",}
    
	\textit{"OPTION": "(A) Fish tank}\\
\textit{(B) Tissue}\\
\textit{(C) Eye}\\
\textit{(D) Water bottle"}\\
\textit{\}}\\
\textit{OUTPUT: C}\\
\\
\textit{Example 2:}
\textit{INPUT: \{}

	\textit{"IMAGE": ,}
    
	\textit{"QUESTION": Carefully observe the shape of the lake in the image. What does it remind you of using your associative thinking? Please choose the best option from the following choices to answer the question.",}
    
	\textit{"OPTION": "(A) Cat}\\
\textit{(B) Phone}\\
\textit{(C) Rock}\\
\textit{(D) Alarm clock"}\\
\textit{\}}\\
\textit{OUTPUT: A}\\
\\
\textit{Referencing the above examples, based on the following latest INPUT information, use associative thinking to analyze the given image, and select the best option from the provided choices to answer the question. Strictly output the result in the same format as shown in the examples.}\\
\\
\textit{INPUT: \{ }

	\textit{"IMAGE": ,}
    
	\textit{"QUESTION": "<question>",}
    
	\textit{"OPTION": "<option>"}\\
\textit{\}}\\
\textit{OUTPUT: }

\end{document}